\documentclass{article}
\usepackage{ijcai24}

\usepackage{times}

\usepackage{soul}
\usepackage{url}
\usepackage[hidelinks]{hyperref}
\usepackage[utf8]{inputenc}
\usepackage[small]{caption}
\usepackage{graphicx}
\usepackage{amsmath}
\usepackage{amsthm}
\usepackage{subcaption}
\usepackage{booktabs}
\usepackage{algorithm}
\usepackage{amsthm}
\usepackage{algorithmic}
\usepackage{setspace}
\usepackage{bm}
\usepackage{multirow}
\usepackage{pifont}
\usepackage{balance}
\usepackage{makecell}
\usepackage{colortbl} 
\usepackage{cuted} 
\definecolor{mygray}{gray}{.9}
\usepackage{hyperref}
\usepackage[switch]{lineno}

\newtheorem{theorem}{Theorem}
\newtheorem{assumption}{Assumption}

\title{Boosting Few-Shot Semantic Segmentation Via Segment Anything Model}

\author{
Chen-Bin Feng$^1$
\and
Qi Lai$^2$\and
Kangdao Liu$^1$\and
Houcheng Su$^1$\and
Chi-Man Vong$^1$
\affiliations
$^1$University of Macau\\
$^2$Shenzhen Institute of Advanced Technology, Chinese Academy of Science\\
\emails
{fengchenbinjacob}@gmail.com
}

\begin{document}

\maketitle

\begin{abstract}
In semantic segmentation, accurate prediction masks are crucial for downstream tasks such as medical image analysis and image editing. Due to the lack of annotated data, few-shot semantic segmentation (FSS) performs poorly in predicting masks with precise contours. Recently, we have noticed that the large foundation model segment anything model (SAM) performs well in processing detailed features. Inspired by SAM, we propose FSS-SAM to boost FSS methods by addressing the issue of inaccurate contour. The FSS-SAM is training-free. It works as a post-processing tool for any FSS methods and can improve the accuracy of predicted masks. Specifically, we use predicted masks from FSS methods to generate prompts and then use SAM to predict new masks. To avoid predicting wrong masks with SAM, we propose a prediction result selection (PRS) algorithm. The algorithm can remarkably decrease wrong predictions. Experiment results on public datasets show that our method is superior to base FSS methods in both quantitative and qualitative aspects.

\end{abstract}

\section{Introduction}
Due to the high cost of manual annotation for semantic segmentation, few-shot semantic segmentation (FSS) \cite{shaban2017one,dong2018few} was proposed. It can segment novel classes in inference time with a few annotated images. The novel classes are unseen during training. It is very useful to segment rare classes with a few annotations in real-world applications. Therefore, FSS can be a supplement for original segmentation methods which are trained to segment common classes.

\begin{figure}
\centering
	\begin{subfigure}{0.24\linewidth}
		\centering
		\includegraphics[height=1.7cm]{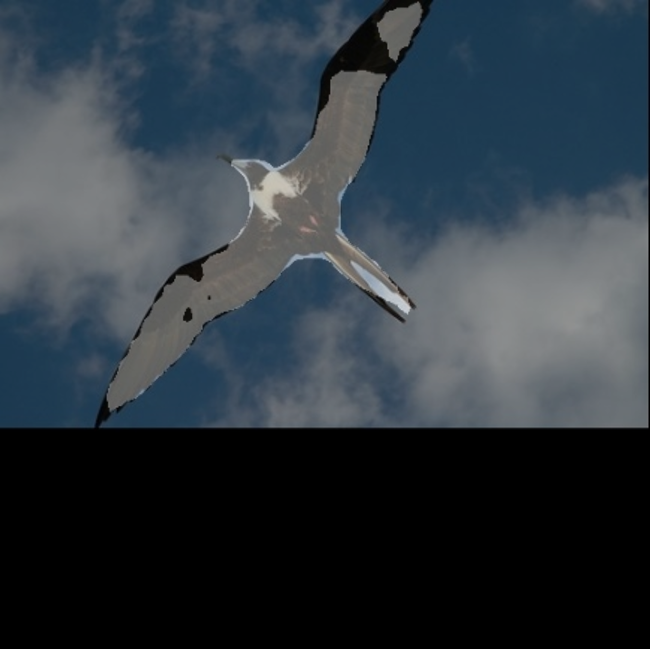}
		\caption{FSS }
	\end{subfigure}
	\centering
 \begin{subfigure}{0.24\linewidth}
		\centering
		\includegraphics[height=1.7cm]{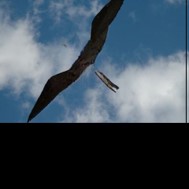}
		\caption{FSS (IP)}
	\end{subfigure}
 	\begin{subfigure}{0.24\linewidth}
		\centering
		\includegraphics[height=1.7cm]{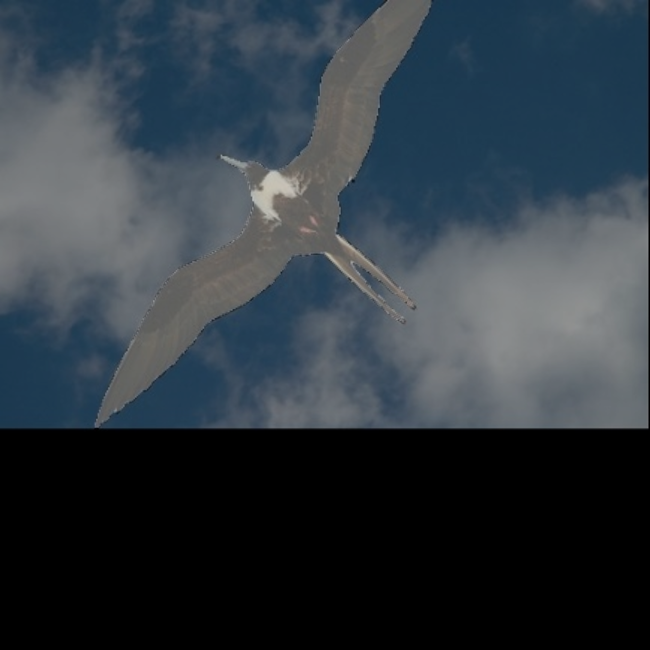}
		\caption{ours}
	\end{subfigure}
	\centering
 \begin{subfigure}{0.24\linewidth}
		\centering
		\includegraphics[height=1.7cm]{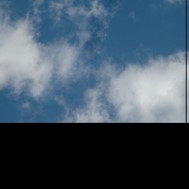}
		\caption{ours (IP)}
	\end{subfigure}
        \vspace{5mm}
        \caption{The qualitative comparison of the segmentation mask (mark in white) and corresponding inpainting (IP) results by the base FSS method and our boosted FSS-SAM.}
        \label{fig:tou}
 
\end{figure}

One of the problems of existing FSS methods is the predicted masks often do not have a clear contour as shown in Figure \ref{fig:tou}. This may be due to the limited number of novel class annotated images, which makes it hard for models to fully learn the detailed edge features of the class. This weakness limits the use of FSS methods in a lot of downstream applications such as medical image analysis \cite{chen2018drinet,zhang2023input}, image inpainting \cite{chu2023rethinking,ko2023continuously}, image composition \cite{chen2023anydoor}, etc.

\begin{figure}
    \centering
    \includegraphics[width=7cm]{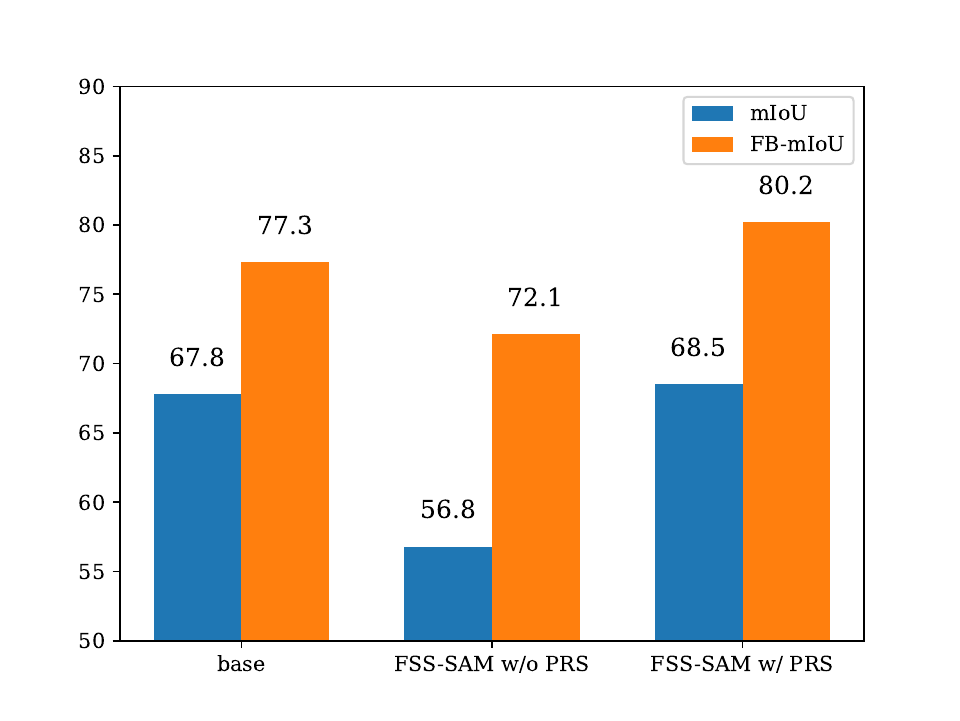}
    \caption{The quantitative comparisons of base FSS, FSS-SAM without PRS algorithm, and FSS-SAM with PRS algorithm in terms of mIoU and FB-mIoU on PASCAL-$5^i$.}
    \label{fig:tou2}
\end{figure}
Recently, a large vision model (LVM) segment anything model (SAM) \cite{kirillov2023segment} has brought tremendous change to semantic segmentation. We consider that it has the potential to solve the problems in FSS. SAM uses 1 billion masks and 11 million images for training, making it have extraordinary zero-shot generalization power, especially in solving detailed features. Therefore, we plan to bridge SAM and FSS to make the predicted mask more accurate. 

An intuitive way to bridge SAM and FSS is to re-implement SAM in the FSS paradigm and finetune on the FSS datasets. However, it will consume a lot of computing resources. Unlike re-implement and finetune, we propose a training-free method that can improve the performance of FSS methods efficiently. This method does not need any extra training process and can be easily plugged into any FSS method. Specifically, we use the prompt engineering approach of SAM to implement a post-processing framework for FSS. We take the output prediction masks of an FSS method to generate prompts. The prompts indicate the rough location of the objects. After that, we input the prompts and the input image to the SAM to get a more accurate new prediction mask. A comparison example can be found in Figure \ref{fig:tou}. The mask prediction result of our FSS-SAM is more accurate. The downstream inpainting result of our method is successful while the base FSS is unsuccessful. It demonstrates the importance of precise masks in downstream tasks.

Our method is useful for predicting more accurate masks. However, as shown in the bar chart of Figure \ref{fig:tou2}, the overall prediction performance of the method is not very well. It is due to the complexity of the scenes for segmentation and the rough prompts that could produce wrong predictions. Therefore, we propose a prediction selection algorithm that can exclude most of the wrong predictions. The principle of this algorithm is to exclude new masks that differ significantly from the original masks. We can see from Figure \ref{fig:tou2} that this algorithm can significantly improve the overall performance and outperform the original FSS method in terms of mIoU and FB-mIoU. The major contributions of this paper can be concluded as follows:
\begin{itemize}
    \item To the best of our knowledge, this is the first work to improve few-shot semantic segmentation in a training-agnostic manner using a large vision model (LVM).
    \item We propose FSS-SAM framework using prompt engineering and a selection algorithm that can exclude wrong predictions.
    \item We evaluate our model by plugging our framework into a state-of-the-art FSS method. The experimental results show that the method combined with our framework is superior to the original FSS method.

\end{itemize}





\section{Related Works}
\subsection{Few-Shot Semantic Segmentation}
Recent FSS methods focus on how to design more advanced model structures to fully utilize the support information to segment query images. There are roughly two types of FSS methods, prototype-based and pixel-based. The prototype-base models transform the features extracted from support images and support masks to some prototypes, then segment query images by matching the prototypes and the query feature maps with cosine similarity or concatenation. For the recent prototype-based methods, NTRENet \cite{liu2022learning} explicitly extracts background prototypes to eliminate wrong matching between support background features and query features. DPCN \cite{liu2022dynamic} uses dynamic convolutional kernels to extract rich support foreground features as prototypes. For pixel-based methods, support and query matching are in a pixel-to-pixel manner. For the recent works, HSNet uses the 4D convolution to do dense feature matching. SCCAN \cite{xu2023self} uses self-calibrated cross-attention and moving window attention to do feature matching. However, these methods pay more attention to the similarity between support features and query features and they do not fully utilize the context information of query images themselves. Therefore, FSS-SAM can help make up for this deficiency.
\subsection{Prompt-Based Large Vision Model}
With the development of large language models (LLM), prompt engineering has become a popular research topic. Prompt engineering can connect pretrained large models and downstream tasks. This enables us to have the potential to use large-scale models to develop more applications. This characteristic is also applicable in the field of large vision models. As an LVM model for segmentation, SAM \cite{kirillov2023segment} has two work modes: automatic and prompt. The automatic mode segments images without other input. We focus on prompt mode. Recent prompt-based applications of SAM include medical image analysis \cite{wu2023medical,zhou2023can}, video object tracking \cite{yang2023track,cheng2023segment}, image editing \cite{yu2023inpaint,chen2023anydoor}, and so on. MSA \cite{wu2023medical} uses an adaptor to fine-tune SAM, aiming to transfer domain knowledge of medical image segmentation to SAM. SAM-Track \cite{cheng2023segment,yu2023inpaint} is a unified video segmentation and tracking framework. It provides two modes: iterative and automatic, Iterative mode enables users to choose objects in multimodal ways. Automatic mode tracks new objects in the subsequent frames. Anydoor \cite{chen2023anydoor} uses SAM as an object segmentation tool for the subsequent image foreground-background composition task. Few-shot semantic segmentation is also a potential downstream task that is suitable for prompt-based large vision models, which is a motivation for our work.

\section{Problem Definition}
For a typical meta-learning paradigm of the few-shot semantic segmentation 
task, each episode includes a query image set $Q$ = $\{$$I^q, M^q$$\}$ and a
support image set S = $\{$$I^S_k, M^S_k$$\}^K_{k=1}$. The episodes are 
randomly sampled in the base image set. The $I^q$ and $M^q$ stand for the 
query images and query masks. The $I^S$ and $M^S$ stand for the support images and support masks. K means the number of support images and masks we 
use to predict a mask for a query image. During inference, the $I^q$, $I^S$, 
and $M^S$ are given. The $M^q$ is the prediction target. The ground truth of the query is $GT^q$.

Our FSS-SAM is an inference-time post-processing method for ordinary FSS methods. Our input is $Q$ = $\{I^q, M^q_{FSS}\}$, where $I^q$ is the query image and $M^q_{FSS}$ is 
the mask predicted by an arbitrary pretrained FSS method. We use $Q$ to predict $Q'$ = $\{M^q_{SAM}\}$,
where $M^q_{SAM}$ is the new predicted mask boosted by SAM.

\begin{figure*}
	\centering
	\includegraphics[width=0.85\textwidth]{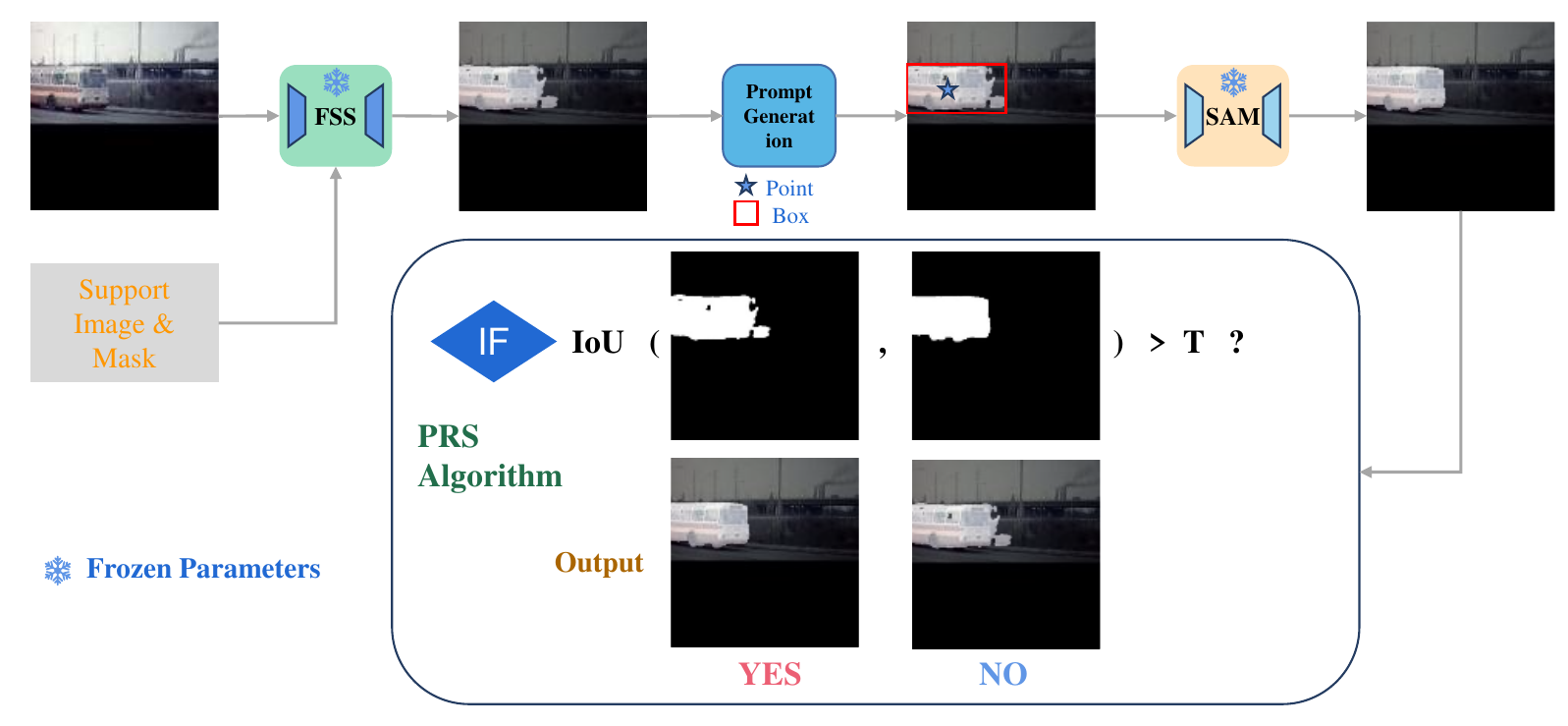}
	\caption{Framework of our FSS-SAM. The FSS model and SAM are all frozen, the parameters are all fixed. In the PRS algorithm, IoU means intersection over union, and T denotes the threshold. The YES and NO indicate whether the condition is met.}
	\label{fig:bigmap}
	\vspace{0.5cm}
\end{figure*}
\section{Method}
\subsection{Framework}
The framework is illustrated in Figure \ref{fig:bigmap}. The framework is in inference mode. We have a query image $I^q$, support image $I^s$, and support mask $M^s$. Firstly, we use a pretrianed FSS method to predict a coarse mask $M^q_{FSS}$. Then we use $M^q_{FSS}$ to generate point and box prompts. The input of SAM is the prompts together with query image $I^q$. Finally, we process inference on pretrianed SAM to get the fine mask $M^q_{SAM}$. After we get prediction results from SAM. We propose a select algorithm named the PRS algorithm. The PRS algorithm can choose better predictions. We illustrate the algorithm in section 4.3.

\begin{figure}

  \includegraphics[width=0.45\textwidth]{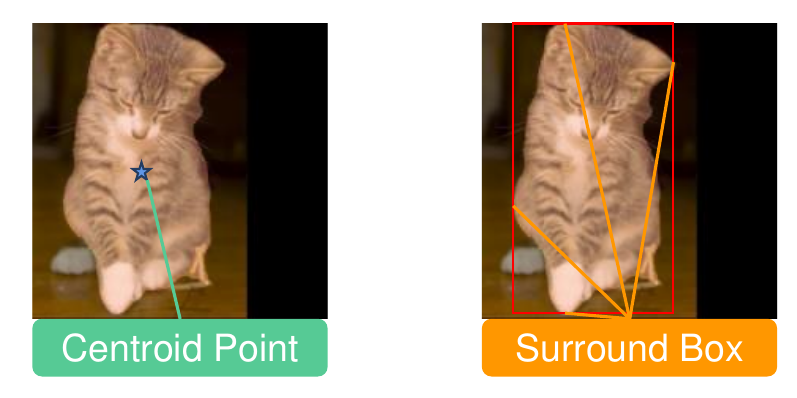}
  \caption{The illustration of different prompts.}
  \label{fig:cat} 

\end{figure}

\subsection{Prompt Generation}
SAM receives images and prompts as input. 
\begin{equation}
    M^q_{SAM} = SAM(I^q,Prompts)
\end{equation}
We need to design suitable prompts to provide SAM with the location of the target objects. SAM receives two kinds of prompts, box and point. So we generate the box and point from the $M^q_{FSS}$. We design ways to get these prompts as illustrated in Figure \ref{fig:cat}.

\textbf{Point} The point we use is the centroid point of the foreground in the $M^q_{FSS}$. We use the moment theory \cite{mukundan1998moment} for calculation. The implementation is based on the OpenCV library \cite{bradski2000opencv}. 


\textbf{Box} The box we use is the largest surrounded box of the foreground area of the $M^q_{FSS}$. The calculation method is as follows:
\begin{equation}
\begin{split}
    x_1 = min(F), y_1 = min(F) \\
    x_2 = max(F), y_2 = max(F)
\end{split}
\end{equation}
where F is the foreground of the $M^q_{FSS}$, $min$ and $max$ is the minimal and maximal value of the foreground, $(x_1,y_1)$ and $(x_2,y_2)$ are the vertex coordinates of the box.

\begin{figure}

  \includegraphics[width=8.5cm]{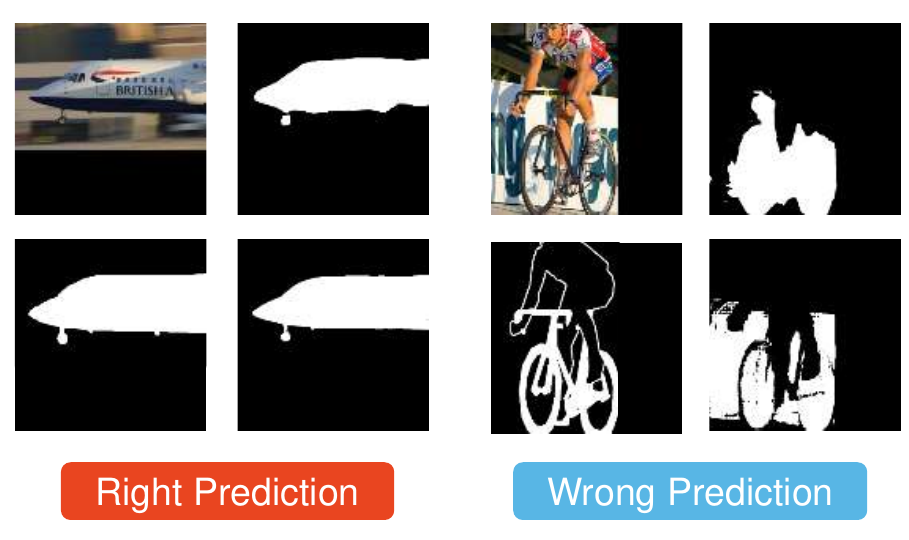}
  \caption{The right and wrong prediction by FSS-SAM. For each sub-figures, The top left is $I^q$, top right is $M^q_{FSS}$, bottom left is $GT^q$, bottom right is $M^q_{SAM}$.}
  \label{fig:rightwrong} 

\end{figure}

\subsection{PRS Algorithm}
\subsubsection{Motivations}
Due to the complexity of image content, especially for some images with multiple objects, not all the boosted FSS-SAM results are right. Illustrations of right and wrong examples are shown in Figure \ref{fig:rightwrong}. The wrong prediction is caused by multiple objects in the scene. The wrong prediction will decrease model performance. To prevent performance from decreasing, we analyze our results. The performance of our method is related to two factors: the performance of the base FSS method, and the performance of our FSS-SAM method. Concretely, there are three situations.
(1) FSS predicts right, and SAM predicts right. The prediction performance of $M^q_{SAM}$ is better than $M^q_{FSS}$ in this situation.
(2) FSS predicts right, and SAM predicts wrong. At this time, $M^q_{SAM}$ is worse than $M^q_{FSS}$. This is the situation that we want to reduce.
(3) FSS predicts wrong. The FSS predicts the wrong locations of the target object. At this time, regardless of whether FSS-SAM predicts the correct object or not, the overall predictive performance cannot be predicted. So, there is no practical value. Based on the three situations, we aim to improve overall performance by decreasing the situation (2).

\subsubsection{Algorithm}
We propose a strategy to reduce the situation (2). Specifically, we want to select better predicting results between $M^q_{FSS}$ and $M^q_{SAM}$. In other words, we want to exclude wrong predictions by SAM. To achieve this goal, we need a judging criterion that helps us to choose $M^q_{FSS}$ or $M^q_{SAM}$ as the final prediction. In order to formulate the judging criteria, we make an assumption:

\begin{assumption}
	\label{ass:1}
	$M^q_{FSS}$ is a coarse mask that gives the right location of the target objects.
\end{assumption}

Based on Assumption \ref{ass:1}, the prompt generated by $M^q_{FSS}$ provides the correct location to SAM, and we can get a valid SAM-boosted mask. 
When the Assumption \ref{ass:1} holds, our method helps to improve performance. For exceptions to this assumption, $M^q_{FSS}$ gives the wrong location. We input prompts at the wrong locations to SAM so that the SAM-boosted masks are wrong either. At this time, neither choosing $M^q_{FSS}$ nor $M^q_{SAM}$ can increase overall performance. So we ignore these cases. 

Based on experimental experience, we find that when a $M^q_{SAM}$ shares a large number of pixels with the corresponding $M^q_{FSS}$, it is likely to be a right prediction and vice versa. Figure \ref{fig:rightwrong} is an example. So we introduce Condition 1: The intersection over union (IoU) of $M^q_{SAM}$ and $M^q_{FSS}$ is greater than a predefined threshold. At this time, we have an empirical theorem: 

\begin{theorem}
	\label{the:1}
	When the Condition 1 meets, the $M^q_{SAM}$ is a right prediction. 
\end{theorem}
Furthermore, we have another empirical theorem that comes from the experiment results:

\begin{theorem}
	\label{the:2}
	When the $M^q_{SAM}$ is a right prediction, it is more accurate than the corresponding $M^q_{FSS}$. 
\end{theorem}

Therefore, we can choose $M^q_{SAM}$ instead of $M^q_{FSS}$ when the Condition 1 is met. Based on the above analysis, we propose a prediction results selection algorithm, abbreviated as PRS algorithm. The algorithm can perform batch processing and output the final prediction results. The detailed algorithm is illustrated in Algorithm \ref{alg:algorithm}.

\begin{algorithm}[tb]
    \caption{Prediction Results Selection}
    \label{alg:algorithm}
    \textbf{Input}: Initial dataset $D_{FSS}$, $D_{SAM}$ containing FSS generated masks $M^q_{FSS}$ and SAM-boosted masks $M^q_{SAM}$. We denote the $i^{th}$ corresponding $M^q_{FSS}$ and $M^q_{SAM}$ as $FSS_i$ and $SAM_i$. \\
    \textbf{Parameter}: Threshold $T$\\
    \textbf{Output}: Final dataset $D_{FSS-SAM}$
    \begin{algorithmic}[1] 
        \WHILE{$FSS_i$, $SAM_i$ in $D_{FSS}$,$D_{SAM}$}
        \STATE Calculate intersection of $FSS_i$ and $SAM_i$.
        \STATE Calculate union of $FSS_i$ and $SAM_i$.
        \STATE Calculate the intersection over union (IoU).
        \IF {IoU $>$ $T$}
        \STATE Append $SAM_i$ to final dataset $D_{FSS-SAM}$.
        \ELSE
        \STATE Append $FSS_i$ to final dataset $D_{FSS-SAM}$.
        \ENDIF
        \ENDWHILE
        \STATE \textbf{return} $D_{FSS-SAM}$
    \end{algorithmic}
\end{algorithm}


\subsubsection{Performance Analysis}

When it comes to evaluation metrics, we discuss individual data and batch data separately. For individual data, it can be classified into three situations.
\begin{itemize}
    \item PRS algorithm choose $M^q_{SAM}$ instead of $M^q_{FSS}$ and the  segmentation accuracy increase. 
    \item PRS algorithm choose $M^q_{SAM}$ instead of $M^q_{FSS}$ and the  segmentation accuracy decrease.
    \item PRS algorithm choose $M^q_{FSS}$ instead of $M^q_{SAM}$. The performance of this method is the same as that of the basic FSS method.
\end{itemize}

For batch data, the above three situations occur simultaneously. Therefore, for overall metrics such as mIoU and FB-mIoU, the proportion and the increased or decreased value of these three situations affect the final performance comparison. We discuss the performance with FB-mIoU and name it FB-mIoU-S, The formula is as follows:
\begin{equation}
    \textrm{FB-mIoU-S} = \frac{\sum_{i=1}^{S} \sum_{j=1}^{N_i} I_{i,j} }{\sum_{i=1}^{S}\sum_{j=1}^{N_i} U_{i,j}}
\end{equation}
where $S$ is the number of situations,  $N_i$ is the number of samples in each situation, $I_{i,j}$, $U_{i,j}$ are the intersection and union of one $M^q_{SAM}$ and the corresponding $GT^q$, separately. When situation one has a greater impact on FB-mIoU values than situation two, the overall performance of $D_{FSS-SAM}$ is better than $D_{SAM}$, and vice versa. From the experiment results, we evaluate that the PRS algorithm increases the overall performance by decreasing situation two. We show how the PRS algorithm improves the overall performance including mIoU and FB-mIoU in section 5.6.

\section{Experiment}

\subsection{Datasets}
We evaluate the performance of our method on two public datasets, PASCAL-$5^i$ \cite{shaban2017one} and COCO-$20^i$ \cite{nguyen2019feature}. PASCAL-$5^i$ contains 20 classes. It is built from PASCAL VOC 2012 \cite{everingham2010pascal}. COCO-$20^i$ has 80 classes. It is created from MSCOCO \cite{lin2014microsoft}. Following previous works \cite{tian2020prior,liu2020crnet}, we split PASCAL-$5^i$ and COCO-$20^i$ into 4 folds evenly for cross-validation. For each fold, the other three folds are used for training, while this fold itself is used for testing. Following previous work, we randomly sampled 1,000 episodes from PASCAL-$5^i$ and COCO-$20^i$ for testing.
\subsection{Evaluation metrics}
We adopt mean intersection over union (mIoU) and foreground-background mIoU (FB-mIoU) as our evaluation metrics. The mIoU measures the mean IoU for all foreground classes. FB-mIoU ignores the classes and treats them as a single FG class. 

\subsection{Implementation details}
We use the codes and pretrianed weights of the official SAM repository \cite{kirillov2023segment} to implement our model. The model does not require training and can be plugged into all the existing few-shot semantic segmentation methods. We use the open-source code and pretrained weights of BAM \cite{lang2022learning} to evaluate our model. The threshold T in the PRS algorithm is empirically set to 0.75. We choose box prompts for evaluation.
\subsection{Performance Comparisons}

\subsubsection{Quantitative Results}
We compare our model with several state-of-the-art models, including  
SG-One (TCYB'19) \cite{zhang2018sg}, PANet (ICCV'19) \cite{wang2019panet}, FWB (ICCV'19) \cite{wang2019panet}, PFENet (TPAMI'20) \cite{tian2020prior}, PRNet(arxiv'20) \cite{liu2020prototype},HSNet (ICCV'21) \cite{min2021hypercorrelation}, CANet (ICCV'19) \cite{zhang2019canet}, PGNet (ICCV'19) \cite{zhang2019pyramid}, PPNet (ECCV'20) \cite{liu2020part}, CWT (ICCV'21) \cite{lu2021simpler}, CyCTR (NeurIPS'21) \cite{zhang2021few}, DCP (IJCAI'22) \cite{lang2022beyond}, NTRENet (CVPR'22) \cite{liu2022learning}, DCAMA (ECCV'22) \cite{shi2022dense}, MLC (ICCV'21) \cite{yang2021mining}, BAM (CVPR'22) \cite{lang2022learning} and ABCNet (CVPR'23) \cite{wang2023rethinking}. The mIoU comparison results on PASCAL-$5^i$ are shown in Table \ref{tab:bigtable1}. We can see that: (1) Our models outperform the state-of-the-art models separately on two kinds of backbones and 1-shot/5-shot settings. (2) The baseline model that is plugged in by our model outperforms the baseline models. For FB-mIoU, the results of PASCAL-$5^i$ are illustrated in Table \ref{tab:smalltable}. We use the average value of the 4 folds. We can see that the baseline model combined with our model outperforms the baseline models with VGG16 \cite{simonyan2014very} and ResNet50 \cite{he2016deep} separately. For the results of COCO-$20^i$ shown in Table \ref{tab:bigtable2}, we know that our model outperforms most of the state-of-the-art models on different backbones and shot settings. Compared to the baseline model, our model performs relatively better. We can conclude that in terms of mIoU and FB-mIoU, our model evidently outperforms the baseline model and other state-of-the-art models on PASCAL-$5^i$, and marginally outperforms other models on COCO-$20^i$. In addition, according to qualitative results, our method performs better than the results currently displayed.


\begin{table}[t]
    \centering
			\begin{tabular}{p{1.5cm}p{2cm}|p{1.5cm}<{\centering}p{1.5cm}<{\centering}}
				\hline
				\multirow{2}{*}{Backbone} & \multirow{2}{*}{Method} & \multicolumn{2}{c}{FB-IoU (\%)} \\ \cline{3-4} 
				&                         & 1-shot         & 5-shot         \\ \hline
				\multirow{4}{*}{VGG16}    
				& PFENet                 & 72.0          & 72.3          \\
				& HSNet                  & 73.4         & 76.6          \\ \cline{2-4}
                & BAM                 & 77.3          & 81.1          \\
				& \cellcolor{mygray}BAM + ours              & \cellcolor{mygray}\textbf{78.2}          & \cellcolor{mygray}\textbf{82.0}          \\ \hline
				\multirow{6}{*}{ResNet50} & PGNet                   & 69.9          & 70.5          \\
				& PPNet                   & 69.2          & 75.8          \\
				& PFENet                  & 73.3          & 73.9          \\
				& HSNet                   & 76.7          & 80.6          \\ \cline{2-4}
                & BAM                   & 79.7          & 82.2          \\
				& \cellcolor{mygray}BAM + ours              & \cellcolor{mygray}\textbf{80.2}          & \cellcolor{mygray}\textbf{82.7}          \\ \hline
		\end{tabular}
    \caption{Performance comparison on PASCAL-$5^i$ in terms of FB-mIoU.}
    \label{tab:smalltable}
\end{table}

\begin{table*}[t]
	\centering
	\resizebox{\linewidth}{!}{
		\renewcommand{\arraystretch}{1.1}
		\begin{tabular}{p{1.3cm}>{\hfill}p{3.7cm}|p{1.0cm}<{\centering}p{1.0cm}<{\centering}p{1.0cm}<{\centering}p{1.0cm}<{\centering}p{1.0cm}<{\centering}|p{1.0cm}<{\centering}p{1.0cm}<{\centering}p{1.0cm}<{\centering}p{1.0cm}<{\centering}p{1.0cm}<{\centering}}
			\hline
			\multirow{2}{*}{Backbone} & \multirow{2}{*}{Method} & \multicolumn{5}{c|}{1-shot}           & \multicolumn{5}{c}{5-shot}            \\ \cline{3-12} 
			&            & Fold-0 & Fold-1 & Fold-2 & Fold-3 & Mean  & Fold-0 & Fold-1 & Fold-2 & Fold-3 & Mean  \\ \hline
			\multirow{7}{*}{VGG16}    & SG-One                  & 40.2 & 58.4 & 48.4 & 38.4 & 46.3 & 41.9 & 58.6 & 48.6 & 39.4 & 47.1 \\
			& PANet      & 42.3  & 58.0  & 51.1  & 41.2  & 48.1 & 51.8  & 64.6  & 59.8  & 46.5  & 55.7 \\
			& FWB        & 47.0  & 59.6  & 52.6  & 48.3  & 51.9 & 50.9  & 62.9  & 56.5  & 50.1  & 55.1 \\
			& PFENet     & 56.9  & 68.2  & 54.4  & 52.4  & 58.0 & 59.0  & 69.1  & 54.8  & 52.9  & 59.0 \\
			& HSNet      & 59.6  & 65.7  & 59.6  & 54.0  & 59.7 & 64.9  & 69.0  & 64.1  & 58.6  & 64.1 \\ \cline{2-12} 
			& BAM    & 63.2  & 70.8  & 66.1  & 57.5  & 64.4 & 67.4  & 73.1  & 70.6  & 64.0  & 68.8 \\
                                            
			& \cellcolor{mygray}BAM + FSS-SAM (ours) & \cellcolor{mygray}\textbf{65.3}  & \cellcolor{mygray}\textbf{72.1}  & \cellcolor{mygray}\textbf{67.6}  & \cellcolor{mygray}\textbf{58.6}  & \cellcolor{mygray}\textbf{65.9} & \cellcolor{mygray}\textbf{69.8}  & \cellcolor{mygray}\textbf{73.9}  & \cellcolor{mygray}\textbf{71.7}  & \cellcolor{mygray}\textbf{65.2}  & \cellcolor{mygray}\textbf{70.2} \\ \hline
			\multirow{13}{*}{ResNet50} & CANet                & 52.5 & 65.9 & 51.3 & 51.9 & 55.4 & 55.5 & 67.8 & 51.9 & 53.2 & 57.1 \\
			& PGNet     & 56.0  & 66.9  & 50.6  & 50.4  & 56.0 & 57.7  & 68.7  & 52.9  & 54.6  & 58.5 \\
			& PPNet     & 48.6  & 60.6  & 55.7  & 46.5  & 52.8 & 58.9  & 68.3  & 66.8  & 58.0  & 63.0 \\
   			& PFENet    & 61.7  & 69.5  & 55.4  & 56.3  & 60.8 & 63.1  & 70.7  & 55.8  & 57.9  & 61.9 \\
   			& CWT    & 56.3      & 62.0      & 59.9      & 47.2      & 56.4 & 61.3     & 68.5      & 68.5      & 56.6      & 63.7 \\
            &CyCTR &65.7 &71.0 &59.5 &59.7 &64.0 &69.3 &73.5 &63.8 &63.5&67.5 \\
			& HSNet      & 64.3  & 70.7  & 60.3  &  60.5  & 64.0 &  70.3  &  73.2  & 67.4  &  67.1  &  69.5 \\ 
   			& NTRENet      & 65.4  & 72.3  & 59.4  &  59.8  & 64.2 &  66.2  &  72.8  & 61.7  &  62.2  &  65.7 \\
           &DCP &63.8 &70.5 &61.2 &55.7 &62.8 &67.2 &73.2 &66.4 &64.5 &67.8 \\
   			& DCAMA     & 66.1  & 71.9  & 59.7  & 57.5  & 63.8 & 70.7  & 72.9  & 63.0  & 65.0  &  67.9 \\ 
            & ABCNet   &  68.8  &  73.4  &  62.3  & 59.5  &  66.0 & 71.7  & 74.2  &  65.4  & 67.0  & 69.6 \\\cline{2-12} 
                     
   			& BAM    &  69.0  &  73.6  &  67.6  & 61.1  &  67.8 & 70.6  & 75.1  &  70.8  & 67.2  & 70.9 \\
			& \cellcolor{mygray}BAM + FSS-SAM (ours) & \cellcolor{mygray}\textbf{70.2}  & \cellcolor{mygray}\textbf{74.0}  & \cellcolor{mygray}\textbf{67.9}  & \cellcolor{mygray}\textbf{62.0}  & \cellcolor{mygray}\textbf{68.5} & 
            \cellcolor{mygray}\textbf{71.9} & 
            \cellcolor{mygray}\textbf{75.3}  & \cellcolor{mygray}\textbf{71.1}  & \cellcolor{mygray}\textbf{68.3}  & \cellcolor{mygray}\textbf{71.6} \\

            \hline
	\end{tabular}}
	\caption{Performance comparison on PASCAL-$5^i$ in terms of mIoU. Results in \textbf{bold} denote the best performance.}
	\label{tab:bigtable1}
\end{table*}

\begin{table*}[t]
	\vspace{0.5cm}
	\centering
	\resizebox{\linewidth}{!}{
		\renewcommand{\arraystretch}{1.1}
		\begin{tabular}{p{1.3cm}>{\hfill}p{3.7cm}|p{1.0cm}<{\centering}p{1.0cm}<{\centering}p{1.0cm}<{\centering}p{1.0cm}<{\centering}p{1.0cm}<{\centering}|p{1.0cm}<{\centering}p{1.0cm}<{\centering}p{1.0cm}<{\centering}p{1.0cm}<{\centering}p{1.0cm}<{\centering}}
			\hline
			\multirow{2}{*}{Backbone} & \multirow{2}{*}{Method} & \multicolumn{5}{c|}{1-shot}           & \multicolumn{5}{c}{5-shot}            \\ \cline{3-12} 
			&            & Fold-0 & Fold-1 & Fold-2 & Fold-3 & Mean  & Fold-0 & Fold-1 & Fold-2 & Fold-3 & Mean  \\ \hline
			\multirow{5}{*}{VGG16}
			& FWB        & 18.4  & 16.7  & 19.6  & 25.4  & 20.0 & 20.9  & 19.2  & 21.9  & 28.4  & 22.6 \\
            & PRNet      & 27.5  & 33.0  & 26.7  & 29.0  & 29.0 & 31.2  & 36.5  & 31.5  & 32.0  & 32.8 \\ 
			& PFENet     & 35.4  & 38.1  & 36.8  & 34.7  & 36.3 & 38.2  & 42.5  & 41.8  & 38.9  & 40.4 \\ \cline{2-12}
            
			& BAM    & 37.5  & 47.0  & 46.4  & \textbf{41.6}  & 43.1 & 48.2  &  52.6  & \textbf{48.6}  & \textbf{48.8}  & \textbf{49.5} \\
			& \cellcolor{mygray}BAM + FSS-SAM (ours) & \cellcolor{mygray}\textbf{37.7}  & \cellcolor{mygray}\textbf{47.6}  & \cellcolor{mygray}\textbf{46.5}  & \cellcolor{mygray}41.5  & \cellcolor{mygray}\textbf{43.3} & \cellcolor{mygray}\textbf{48.4}  & \cellcolor{mygray}\textbf{52.8}  & \cellcolor{mygray}48.4  & \cellcolor{mygray}48.7  & \cellcolor{mygray}\textbf{49.5} \\ \hline                                            
      
			\multirow{9}{*}{ResNet50} & CWT                  & 32.2 &  36.0 & 31.6 & 31.6 & 32.9  &  40.1 & 43.8 & 39.0 & 42.4 & 41.3  \\
			& CyCTR      & 38.9  & 43.0  & 39.6  & 39.8  & 40.3 &41.1  & 48.9  & 45.2  & 47.0  & 45.6\\
           
			& HSNet      & 36.7  &41.4  & 39.5  &  39.1  & 39.2 &  44.4  &  49.7  & 46.1  &  45.5  &  46.4 \\ 
   			& NTRENet      & 36.8  & 42.6  & 39.9  &  37.9  & 39.3 &  38.2  &  44.1  & 40.4  &  38.4  &  40.3 \\
      	 & DCP      & 40.9  & 43.8  & 42.6  & 38.3   & 41.4 & 45.8  & 49.7  & 43.7  & 46.6  & 46.5 \\
   			& DCAMA     & 41.9  & 45.1  & 44.4  & 41.7  & 43.3 & 45.9  & 50.5  &50.7  & 46.0  &   48.3 \\
              & ABCNet     & \textbf{42.3}  & 46.2  & 46.0  & 42.0 & 44.1& 45.5  & 51.7  &52.6   & 46.4  &   49.1 \\ \cline{2-12}
                      
   			& BAM   &  39.0  &  \textbf{50.6}  &  47.5  & \textbf{43.4}  &  45.1 & 47.2  & \textbf{54.2}  &  \textbf{48.3}  & \textbf{47.1}  & \textbf{49.2} \\
      
			& \cellcolor{mygray}BAM + FSS-SAM (ours) & \cellcolor{mygray}39.1  & \cellcolor{mygray}50.4  & \cellcolor{mygray}\textbf{48.4}  & \cellcolor{mygray}43.1  & \cellcolor{mygray}\textbf{45.3} & \cellcolor{mygray}\textbf{47.3}  & \cellcolor{mygray}54.1  & \cellcolor{mygray}48.2  & \cellcolor{mygray}\textbf{47.1}  & \cellcolor{mygray}\textbf{49.2} \\   \hline
	\end{tabular}}
	\caption{Performance comparison on COCO-$20^i$ in terms of mIoU. Results in \textbf{bold} denote the best performance.}
	\label{tab:bigtable2}
\end{table*}

\subsubsection{Qualitative Results}
We compare the qualitative results between our method and the baseline FSS method. From Figure \ref{fig:pascal} and Figure \ref{fig:coco}, we can see that our prediction results have more accurate edges and details. Compared to the ground truth masks, our results are more accurate. Therefore, the quantitative results of our method should be increased if the ground truth masks are more accurate.
\subsection{Ablation Studies}
\subsubsection{Different Prompts}
\begin{figure}

  \includegraphics[width=8.5cm]{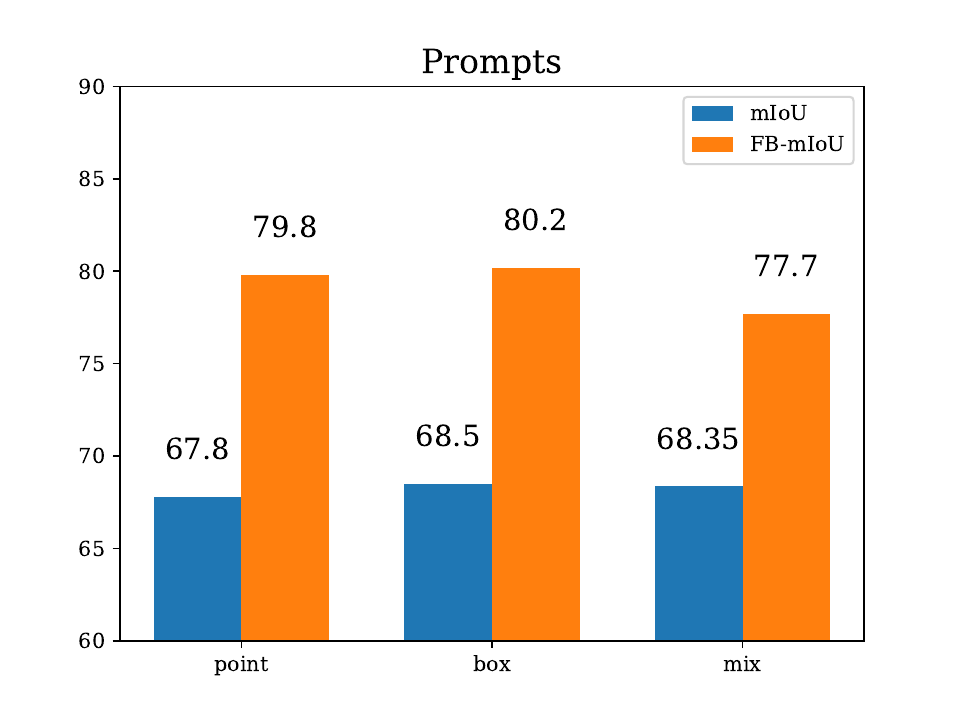}
  \caption{The illustration of performance using different prompts. We set T=0.75 in the PRS algorithm.}
  \label{fig:figure5} 

\end{figure}

\begin{figure*}
\vspace{-1.5cm}
	\centering
	\includegraphics[width=18cm]{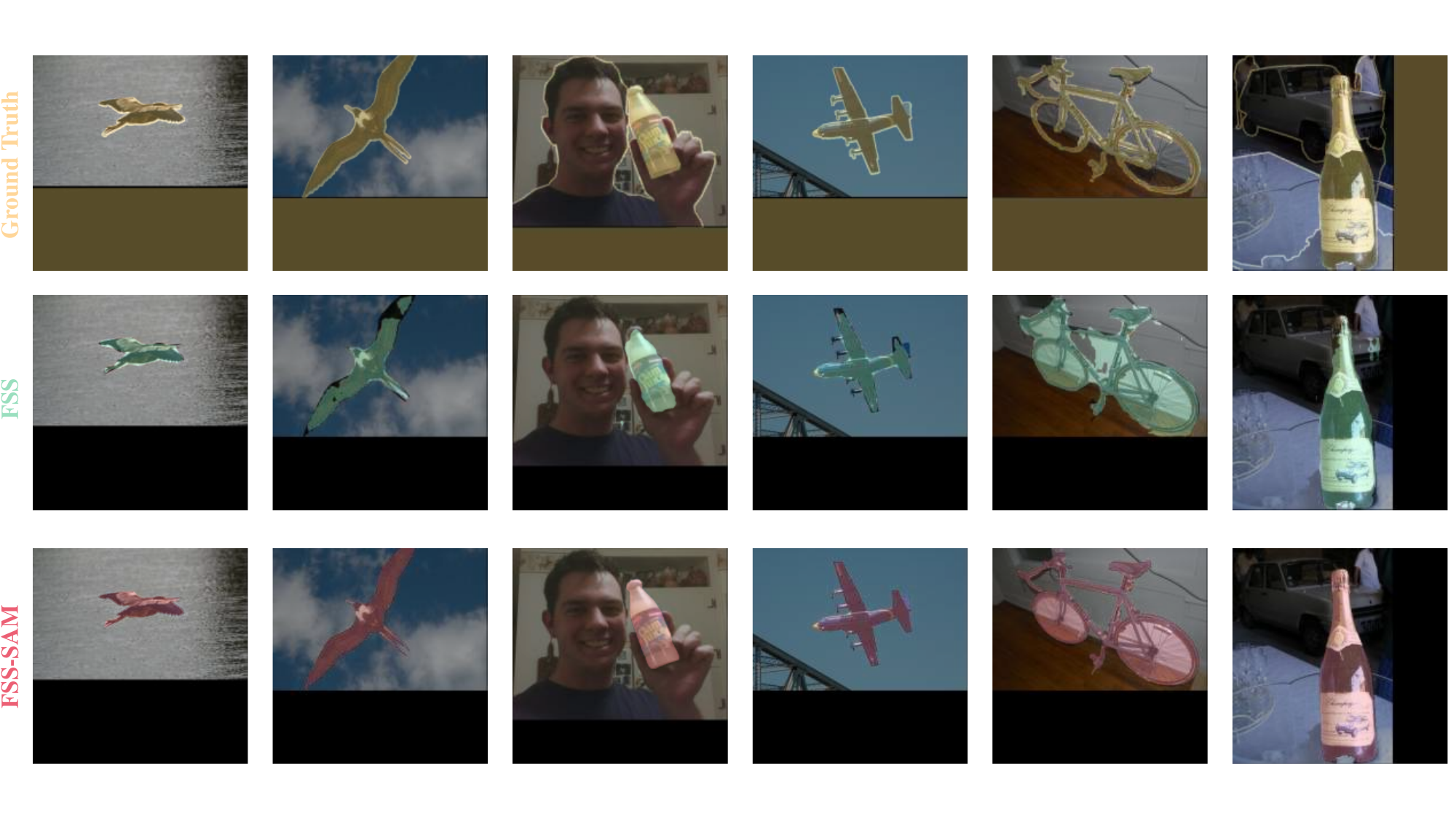}
	\caption{Qualitative comparison results on PASCAL-$5^i$.}
	\label{fig:pascal}
\end{figure*}

\begin{figure*}
	\centering
	\includegraphics[width=18cm]{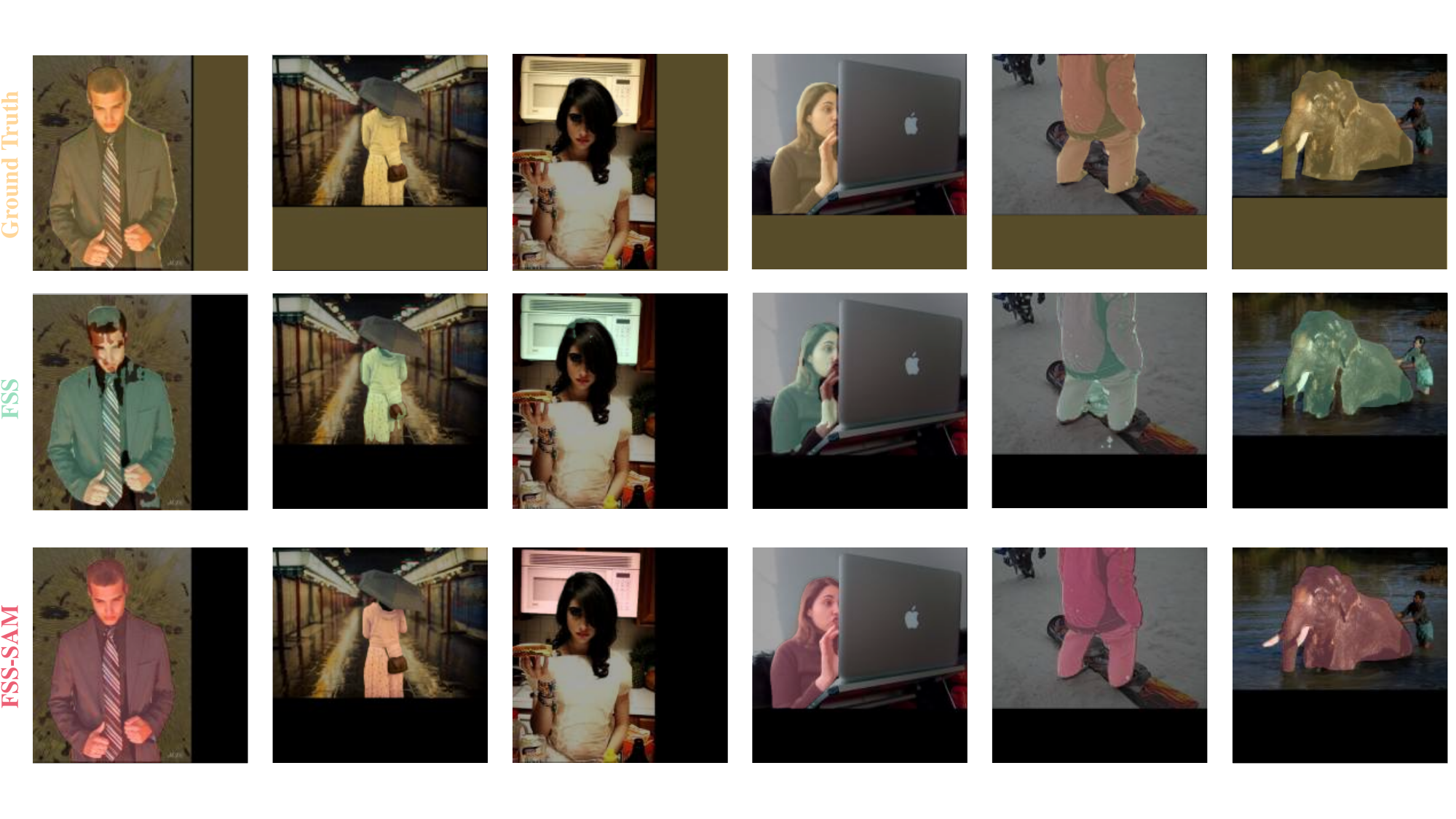}
	\caption{Qualitative comparison results on COCO-$20^i$.}
	\label{fig:coco}
\end{figure*}

We evaluate the performance differences by using different prompts. There are three types of prompts, including box prompts, point prompts, and mixed box and point prompts. The comparison results of ResNet50 backbone, 1-shot setting on PASCAL-$5^i$ are shown in Figure \ref{fig:figure5}. We can see from the figure that the performances of the three kinds of prompts are at the same level. Therefore, for our method, choosing any type of prompt is acceptable. In our experiment, we adopt the box prompt.

\subsection{Parameter Sensitive Analysis}
\subsubsection{T in PRS Algorithm}
\begin{figure}

  \includegraphics[width=8.5cm]{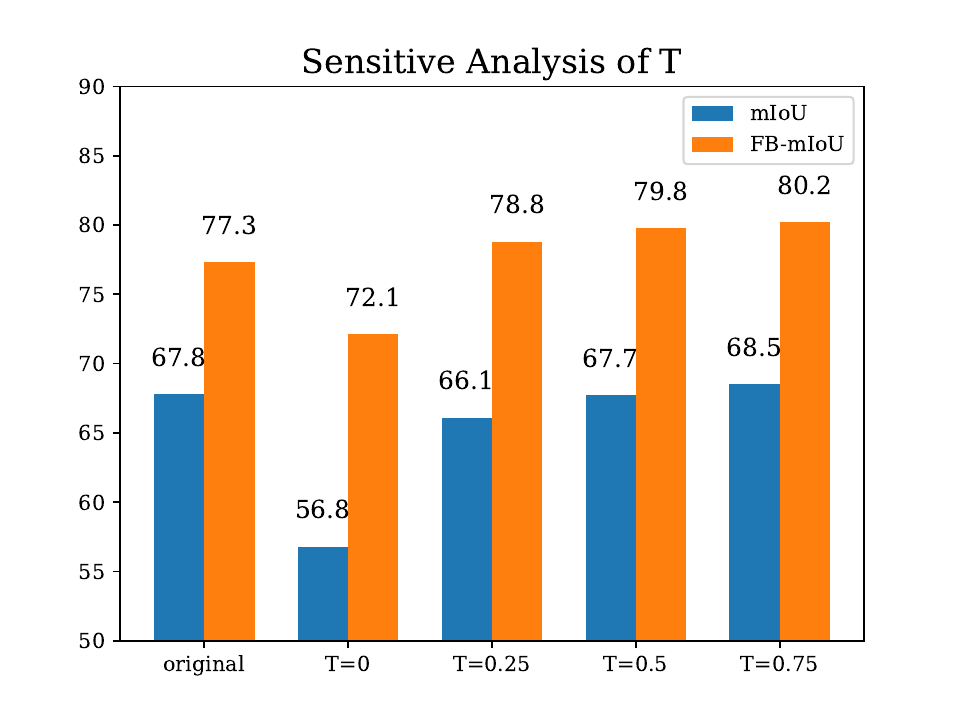}
  \caption{The sensitive analyst of T in PRS algorithm.}
  \label{fig:figure6} 

\end{figure}

Threshold T is the only hyperparameter in the PRS algorithm. It is important to use it to adjust the quantity of boosted FSS-SAM outputs and FSS output. If we set T=1, the PRS algorithm will use the original FSS outputs. On the contrary, if we set T=0, the PSR algorithm will use all the FSS-SAM outputs. When T=0, the model is equivalent to the FSS-SAM without the PRS algorithm. We evaluate the performances that T = 0, 0.25, 0.5, 0.75, 1 under ResNet50 backbone, 1-shot setting on PASCAL-$5^i$. The results are shown in Figure \ref{fig:figure6}. From the figure, we can see that (1) The performance of the T=0 case is lower than the original case. This means using FSS-SAM without the PRS algorithm does not well. (2) The performance of T=0.25, 0.5, and 0.75 continues to improve, and the performance of T=0.75 reaches its best. From the experimental results, we can conclude that the PRS algorithm helps to improve the overall performance of FSS-SAM. Moreover, choosing an appropriate T is important.

\section{Conclusion}
In this paper, we propose a training-free plug-and-play post-processing method that can boost any few-shot semantic segmentation method to get more accurate prediction masks. We leverage the powerful segmentation ability of the large vision model SAM to implement our method. We find that directly using SAM could generate some wrong predictions. Therefore, we propose PRS algorithm to exclude wrong predictions and increase overall prediction performance. Extensive experiments evaluate the superiority of our method.

\end{document}